\documentclass[12pt]{article}

\usepackage{amssymb}
\usepackage{amsmath}
\usepackage[authoryear]{natbib}
\usepackage{geometry}
\usepackage{graphicx}

\renewcommand{\vec}{\mathbf}

\begin{document}

\title{Everything is Connected: Graph Neural Networks}


\author{\Large Petar Veli\v{c}kovi\'{c}}

\date{DeepMind / University of Cambridge}

\maketitle

\begin{abstract}
\begin{center}
\includegraphics[width=\linewidth]{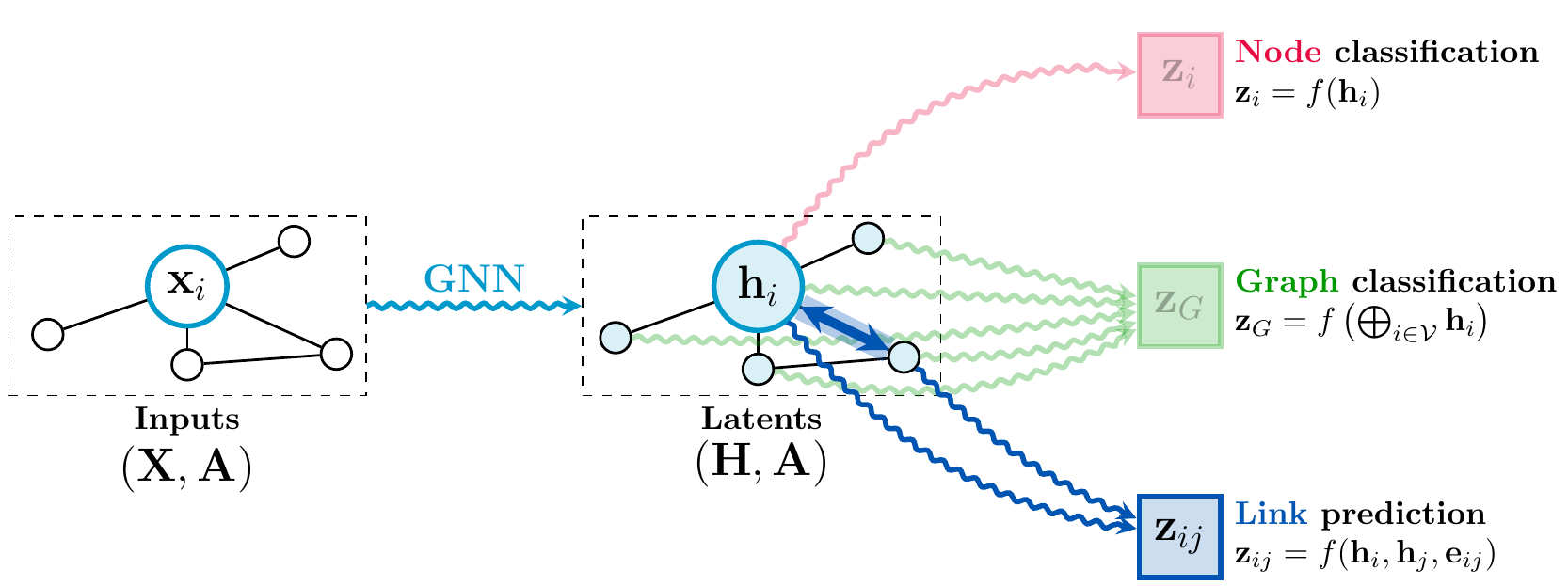}
\end{center} \vspace{1em}
\noindent In many ways, \textbf{graphs} are the main modality of data we receive from \textbf{nature}. This is due to the fact that most of the patterns we see, both in natural and artificial systems, are elegantly representable using the language of graph structures. Prominent examples include molecules (represented as graphs of atoms and bonds), social networks and transportation networks. This potential has already been seen by key scientific and industrial groups, with already-impacted application areas including traffic forecasting, drug discovery, social network analysis and recommender systems. Further, some of the most successful domains of application for machine learning in previous years---images, text and speech processing---can be seen as special cases of graph representation learning, and consequently there has been significant exchange of information between these areas. The main aim of this short survey is to enable the reader to assimilate the key concepts in the area, and position graph representation learning in a proper context with related fields.
\end{abstract}






\section{Introduction: Why study data on graphs?}\label{lab:intro}

In this survey, I will present a vibrant and exciting area of deep learning research: graph representation learning. Or, put simply, building machine learning models over data that lives on \emph{graphs} (interconnected structures of \emph{nodes} connected by \emph{edges}). These models are commonly known as \emph{graph neural networks}, or \textbf{GNNs} for short.

There is very good reason to study data on graphs. From the molecule (a graph of \emph{atoms} connected by chemical \emph{bonds}) all the way to the connectomic structure of the brain (a graph of \emph{neurons} connected by \emph{synapses}), graphs are a universal language for describing living organisms, at all levels of organisation. Similarly, most relevant artificial constructs of interest to humans, from the transportation network (a graph of \emph{intersections} connected by \emph{roads}) to the social network (a graph of \emph{users} connected by \emph{friendship links}), are best reasoned about in terms of graphs.

This potential has been realised in recent years by both scientific and industrial groups, with GNNs now being used to discover novel potent antibiotics \citep{stokes2020deep}, serve estimated travel times in Google Maps \citep{derrow2021eta}, power content recommendations in Pinterest \citep{ying2018graph} and product recommendations in Amazon \citep{hao2020p}, and design the latest generation of machine learning hardware: the TPUv5 \citep{mirhoseini2021graph}. Further, GNN-based systems have helped mathematicians uncover the hidden structure of mathematical objects \citep{davies2021advancing}, leading to new top-tier conjectures in the area of representation theory \citep{blundell2021towards}. It would not be an understatement to say that billions of people are coming into contact with predictions of a GNN, on a day-to-day basis. As such, it is likely a valuable pursuit to study GNNs, even without aiming to directly contribute to their development.

Beyond this, it is likely that the very \emph{cognition} processes driving our reasoning and decision-making are, in some sense, \emph{graph-structured}. That is, paraphrasing a quote from \citet{forrester1971counterintuitive}, nobody really imagines in their head all the information known to them; rather, they imagine only selected \emph{concepts}, and \emph{relationships} between them, and use those to represent the real system. If we subscribe to this interpretation of cognition, it is quite unlikely that we will be able to build a generally intelligent system without some component relying on graph representation learning. Note that this finding does not clash with the fact that many recent skillful ML systems are based on the Transformer architecture \citep{vaswani2017attention}---as we will uncover in this review, Transformers are themselves a special case of GNNs.

\section{The fundamentals: Permutation equivariance and invariance}

In the previous section, we saw \emph{why} it is a good idea to study data that lives on graphs. Now we will see \emph{how to learn} useful functions over graph-structured data. The exposition largely follows \citet{bronstein2021geometric}.

With graph-structured inputs, we typically assume a graph $\mathcal{G}=(\mathcal{V},\mathcal{E})$; that is, we have a set of \emph{edges} $\mathcal{E}\subseteq\mathcal{V}\times\mathcal{V}$, which specifies pairs of nodes in $\mathcal{V}$ that are connected. 

As we are interested in representation learning over the nodes, we attach to each node $u\in\mathcal{V}$ a feature vector, $\vec{x}_u\in\mathbb{R}^k$. The main way in which this data is \emph{presented} to a machine learning model is in the form of a \emph{node feature matrix}. That is, a matrix $\vec{X}\in\mathbb{R}^{|\mathcal{V}|\times k}$ is prepared by stacking these features:
\begin{equation}
    \vec{X}=\left[\vec{x}_1, \vec{x}_2, \dots, \vec{x}_{|\mathcal{V}|}\right]^\top
\end{equation}
that is, the $i$th row of $\vec{X}$ corresponds to $\vec{x}_i$.

There are many ways to represent $\mathcal{E}$; since our context is one of linear algebra, we will use the \emph{adjacency matrix}, $\vec{A}\in\mathbb{R}^{|\mathcal{V}|\times|\mathcal{V}|}$:
\begin{equation}
    a_{uv} = \begin{cases}1 & (u, v)\in\mathcal{E}\\
    0 & (u, v)\notin\mathcal{E}
    \end{cases}
\end{equation}
Note that it is often possible, especially in biochemical inputs, that we want to attach more information to the edges (such as distance scalars, or even entire feature vectors). I deliberately do not consider such cases to retain clarity---the conclusions we make would be the same in those cases.

However, the very act of using the above representations imposes a \emph{node ordering}, and is therefore an arbitrary choice which does not align with the nodes and edges being unordered! Hence, we need to make sure that permuting the nodes and edges ($\vec{P}\vec{A}\vec{P}^\top$, for a permutation matrix $\vec{P}$), does not change the outputs. We recover the following rules a GNN must satisfy:
\begin{align}
    f(\vec{P}\vec{X}, \vec{P}\vec{A}\vec{P}^\top) &= f(\vec{X}, \vec{A}) & \mathrm{(Invariance)}\\
    \vec{F}(\vec{P}\vec{X}, \vec{P}\vec{A}\vec{P}^\top) &= \vec{P}\vec{F}(\vec{X}, \vec{A}) & \mathrm{(Equivariance)}
\end{align}
Here we assumed for simplicity that the functions $f$, $\vec{F}$ do not change the adjacency matrix, so we assume they only return graph or node-level outputs.

Further, the graph's edges allow for a \emph{locality} constraint in these functions. Much like how a CNN operates over a small neighbourhood of each pixel of an image, a GNN can operate over a neighbourhood of a node. One standard way to define this neighbourhood, $\mathcal{N}_u$, is as follows:
\begin{equation}
    \mathcal{N}_u=\{v\ |\ (u, v)\in\mathcal{E}\vee (v, u)\in\mathcal{E}\}
\end{equation}
Accordingly, we can define the multiset of all neighbourhood features, $\vec{X}_{\mathcal{N}_u}$:
\begin{equation}
    \vec{X}_{\mathcal{N}_u} = \{\!\!\{\vec{x}_v\ |\ v\in\mathcal{N}_u\}\!\!\}
\end{equation}
And our local function, $\phi$, can take into account the neighbourhood; that is:
\begin{equation}
    \vec{h}_u = \phi(\vec{x}_u, \vec{X}_{\mathcal{N}_u})\qquad \qquad \vec{F}(\vec{X}) = \left[\vec{h}_1, \vec{h}_2, \dots, \vec{h}_{|\mathcal{V}|}\right]^\top
\end{equation}
Through simple linear algebra manipulation, it is possible to show that if $\phi$ is permutation invariant in $\vec{X}_{\mathcal{N}_u}$, then $\vec{F}$ will be permutation equivariant. The remaining question is, how do we define $\phi$? 

\section{Graph Neural Networks}

Needless to say, defining $\phi$ is one of the most active areas of machine learning research today. Depending on the literature context, it may be referred to as either ``diffusion'', ``propagation'', or ``message passing''. As claimed by \citet{bronstein2021geometric}, most of them can be classified into one of three spatial flavours:
\begin{align}
    \vec{h}_u &= \phi\left(\vec{x}_u,\bigoplus_{v\in\mathcal{N}_u}c_{vu}\psi(\vec{x}_v)\right) & \mathrm{(Convolutional)}\label{eqn:conv}\\
    \vec{h}_u &= \phi\left(\vec{x}_u,\bigoplus_{v\in\mathcal{N}_u}a(\vec{x}_u, \vec{x}_v)\psi(\vec{x}_v)\right) & \mathrm{(Attentional)}\label{eqn:att}\\
    \vec{h}_u &= \phi\left(\vec{x}_u,\bigoplus_{v\in\mathcal{N}_u}\psi(\vec{x}_u, \vec{x}_v)\right) & \mathrm{(Message\text{-}passing)}\label{eqn:three}
\end{align}
where $\psi$ and $\phi$ are neural networks---e.g. $\psi(\vec{x}) = \mathrm{ReLU}(\vec{W}\vec{x} + \vec{b})$, and $\bigoplus$ is any permutation-invariant aggregator, such as $\sum$, averaging, or $\max$. The expressive power of the GNN progressively increases going from Equation \ref{eqn:conv} to \ref{eqn:three}, at the cost of interpretability, scalability, or learning stability. For most tasks, a careful tradeoff is needed when choosing the right flavour. 

This review does not attempt to be a comprehensive overview of specific GNN layers. That being said: representative \emph{convolutional} GNNs include the Chebyshev network \citep[ChebyNet]{defferrard2016convolutional}, graph convolutional network \citep[GCN]{kipf2017semisupervised} and the simplified graph convolution \citep[SGC]{wu2019simplifying}; representative \emph{attentional} GNNs include the mixture model CNN \citep[MoNet]{monti2017geometric}, graph attention network \citep[GAT]{velickovic2018graph} and its recent ``v2'' variant \citep[GATv2]{brody2022how}; and representative \emph{message-passing} GNNs include interaction networks \citep[IN]{battaglia2016interaction}, message passing neural networks \citep[MPNN]{gilmer2017neural} and graph networks \citep[GN]{battaglia2018relational}.

Given such a GNN layer, we can learn (m)any interesting tasks over a graph, by appropriately combining $\vec{h}_u$. I exemplify the three principal such tasks, grounded in biological examples:

{\bf Node classification.} If the aim is to predict targets for each node $u\in\mathcal{V}$, then our output is equivariant, and we can learn a shared classifier directly on $\vec{h}_u$. A canonical example of this is classifying protein functions (e.g. using gene ontology data \citep{zitnik2017predicting}) in a given protein-protein interaction network, as first done by GraphSAGE \citep{hamilton2017inductive}.

{\bf Graph classification.} If we want to predict targets for the entire graph, then we want an invariant output, hence need to first \emph{reduce} all the $\vec{h}_u$ into a common representation, e.g. by performing $\bigoplus_{u\in\mathcal{V}} \vec{h}_u$, then learning a classifier over the resulting flat vector. A canonical example is classifying molecules for their quantum-chemical properties \citep{gilmer2017neural}, estimating pharmacological properties like toxicity or solubility \citep{duvenaud2015convolutional,xiong2019pushing,jiang2021could} or virtual drug screening \citep{stokes2020deep}.

{\bf Link prediction.} Lastly, we may be interested in predicting properties of \emph{edges} $(u, v)$, or even predicting whether an edge \emph{exists}; giving rise to the name ``link prediction''. In this case, a classifier can be learnt over the concatenation of features $\vec{h}_u\|\vec{h}_v$, along with any given edge-level features. Canonical tasks include predicting links between drugs and diseases---drug repurposing \citep{morselli2021network}, drugs and targets---binding affinity prediction \citep{lim2019predicting,jiang2020drug}, or drugs and drugs---predicting adverse side-effects from polypharmacy \citep{zitnik2018modeling,deac2019drug}.

It is possible to use the building blocks from the principal tasks above to go beyond classifying the entities given by the input graph, and have systems that \emph{produce} novel molecules \citep{mercado2021graph} or even perform \emph{retrosynthesis}---the estimation of which reactions to utilise to synthesise given molecules \citep{somnath2021learning,liu2022retrognn}. 

A natural question arises, following similar discussions over sets \citep{zaheer2017deep,wagstaff2019limitations}: Do GNNs, as given by Equation \ref{eqn:three}, represent \emph{all} of the valid permutation-equivariant functions over graphs? Opinions are divided. Key results in previous years seem to indicate that such models are fundamentally limited in terms of problems they can solve \citep{xu2018how,morris2019weisfeiler}. However, most, if not all, of the proposals for addressing those limitations are still expressible using the pairwise message passing formalism of Equation \ref{eqn:three}; the main requirement is to carefully modify the \emph{graph} over which the equation is applied \citep{velickovic2022message}. To supplement this further, \citet{Loukas2020What} showed that, under proper initial features, sufficient depth-width product (\#layers $\times$ $\dim\vec{h}_u$), and correct choices of $\psi$ and $\phi$, GNNs in Equation \ref{eqn:three} are \emph{Turing universal}---likely to be able to simulate \emph{any} computation which any computer can perform over such inputs.

All points considered, it is the author's opinion that the formalism in this section is likely all we need to build powerful GNNs---although, of course, different perspectives may benefit different problems, and existence of a powerful GNN does not mean it is easy to find using stochastic gradient descent.

\section{GNNs without a graph: Deep Sets and Transformers}

Throughout the prior section, we have made a seemingly innocent assumption: that we are \emph{given} an input graph (through $\vec{A}$). However, very often, not only will there not be a clear choice of $\vec{A}$, but we may not have any prior belief on what $\vec{A}$ even is. Further, even if a ground-truth $\vec{A}$ is given \emph{without noise}, it may not be the optimal \emph{computation graph}: that is, passing messages over it may be problematic---for example, due to bottlenecks \citep{alon2021on}. As such, it is generally a useful pursuit to study GNNs that are capable of modulating the input graph structure.

Accordingly, let us assume we only have a node feature matrix $\vec{X}$, but no adjacency. One simple option is the ``pessimistic'' one: assume there are no edges at all, i.e. $\vec{A}=\vec{I}$, or $\mathcal{N}_u=\{u\}$. Under such an assumption, Equations \ref{eqn:conv}--\ref{eqn:three} \emph{all} reduce to $\vec{h}_u = \phi(\vec{x}_u)$, yielding the Deep Sets model \citep{zaheer2017deep}. Therefore, no power from graph-based modelling is exploited here.

The converse option (the ``lazy'' one) is to, instead, assume a \emph{fully-connected} graph; that is $\vec{A}=\vec{1}\vec{1}^\top$, or $\mathcal{N}_u=\mathcal{V}$. This then gives the GNN the full potential to exploit any edges deemed suitable, and is a very popular choice for smaller numbers of nodes. It can be shown that convolutional GNNs (Equation \ref{eqn:conv}) would still reduce to Deep Sets in this case, which motivates the use of a stronger GNN. The next model in the hierarchy, attentional GNNs (Equation \ref{eqn:att}), reduce to the following equation:
\begin{equation}
    \vec{h}_u = \phi\left(\vec{x}_u, \bigoplus_{v\in\mathcal{V}}a(\vec{x}_u, \vec{x}_v)\psi(\vec{x}_v)\right)
\end{equation}
which is essentially the forward pass of a Transformer \citep{vaswani2017attention}. To reverse-engineer why Transformers appear here, let us consider the NLP perspective. Namely, words in a sentence \emph{interact} (e.g. subject-object, adverb-verb). Further, these interactions are not trivial, and certainly not \emph{sequential}---that is, words can interact even if they are many sentences apart\footnote{This insight may also partly explain why RNNs or CNNs have been seen as suboptimal language models: they implicitly assume only neighbouring words directly interact.}. Hence, we may want to use a \emph{graph} between them. But what \emph{is} this graph? Not even annotators tend to agree, and the optimal graph may well be task-dependant. In such a setting, a common assumption is to use a complete graph, and let the network infer relations by itself---at this point, the Transformer is all but rederived. For an in-depth rederivation, see \citet{joshi2020transformers}.

Another reason why Transformers have become such a dominant GNN variant is the fact that using a fully connected graph structure allows to express all model computations using \emph{dense matrix products}, and hence their computations align very well with current prevalent accelerators (GPUs and TPUs). Further, they have a more favourable storage complexity than the message passing variant (Equation \ref{eqn:three}). Accordingly, Transformers can be seen as GNNs that are currently winning the hardware lottery \citep{hooker2021hardware}!

Before closing this section, it is worth noting a \emph{third} option to learning a GNN without an input graph: to \emph{infer} a graph structure to be used as edges for a GNN. This is an emerging area known as \emph{latent graph inference}. It is typically quite challenging, since edge selection is a non-differentiable operation, and various paradigms have been proposed in recent years to overcome this challenge: nonparametric \citep{wang2019dynamic,deac2022expander}, supervised \citep{velivckovic2020pointer}, variational \citep{kipf2018neural}, reinforcement \citep{kazi2022differentiable} and self-supervised learning \citep{fatemi2021slaps}.

\section{GNNs beyond permutation equivariance: Geometric Graphs}

To conclude our discussion, we revisit another assumption: we have assumed our graphs to be a discrete, unordered, collection of nodes and edges---hence, only susceptible to permutation symmetries. But in many cases, this is not the entire story! The graph, in fact, may often be endowed with some spatial \emph{geometry}, which will be very useful to exploit. Molecules, and their three-dimensional conformer structure, are a classical example of this.

In general, we will assume our inputs are \emph{geometric graphs}: nodes are endowed with both \emph{features}, $\vec{f}_u$, and \emph{coordinates}, $\vec{x}_u\in\mathbb{R}^3$. We may be interested in designing a model that is equivariant not only to permutations, but also 3D rotations, translations and reflections (the Euclidean group, $\mathrm{E}(3)$). 

An $\mathrm{E}(3)$-equivariant message passing layer transforms the coordinates and features, and yields updated features $\vec{f}'_u$ and coordinates $\vec{x}'_u$. There exist many GNN layers that obey $\mathrm{E}(n)$ equivariance, and one particularly elegant solution was proposed by \citet{satorras2021n}:
\begin{align}\label{eqn:en}
    \vec{f}'_u &= \phi\left(\vec{f}_u,\bigoplus_{v\in\mathcal{N}_u}\psi_\mathrm{f}\left(\vec{f}_u, \vec{f}_v, \|\vec{x}_u - \vec{x}_v\|^2\right)\right)\\
    \vec{x}'_u &= \vec{x}_u + \sum_{v\neq u}(\vec{x}_u-\vec{x}_v)\psi_\mathrm{x}\left(\vec{f}_u, \vec{f}_v, \|\vec{x}_u - \vec{x}_v\|^2\right)
\end{align}
The key insight behind this model is that rotating, translating or reflecting coordinates does not change their distances $\|\vec{x}_u-\vec{x}_v\|^2$, i.e., such operations are \emph{isometries}. Hence, if we roto-translate all nodes as $\vec{x}_u\leftarrow\vec{R}\vec{x}_u + \vec{b}$, the output features $\vec{f}'_u$ remain unchanged, while the output coordinates transform accordingly: $\vec{x}'_u\leftarrow\vec{R}\vec{x}'_u + \vec{b}$.

While indeed highly elegant, a model like this hides a caveat: it only works over \emph{scalar} features $\vec{f}_u$. If our model needs to support any kind of \emph{vector} input (e.g. \emph{forces} between atoms), the model in Equation \ref{eqn:en} would not suffice, because the vectors would need to appropriately rotate with $\vec{R}$. \citet{satorras2021n} do propose a variant that allows for explicitly updating vector features, $\vec{v}_u$:
\begin{equation}
    \vec{v}'_u = \phi_\mathrm{v}(\vec{h}_u)\vec{v}_u + C\sum_{v\neq u} (\vec{x}_u - \vec{x}_v)\phi_\mathrm{x}\left(\vec{f}_u, \vec{f}_v, \|\vec{x}_u - \vec{x}_v\|^2\right),\quad
    \vec{x}'_u = \vec{x}_u + \vec{v}'_u
\end{equation}
But these issues will continue to arise, as features get ``tensored up''. Hence, in such circumstances, it might be useful to instead characterise a generic equation that supports \emph{all} possible roto-translation equivariant models, and then learning its parameters. Such an analysis was done in Tensor Field Networks \citep{thomas2018tensor} for point clouds, and then extended to $\mathrm{SE}(3)$-Transformers for general graphs \citep{fuchs2020se}.

Perhaps a fitting conclusion of this survey is a simple realisation: having showed how both Transformers and geometric equivariance constraints play a part within the context of GNNs, we now have all of the key building blocks to define some of the most famous geometric GNN architectures in the wild, such as AlphaFold 2 \citep{jumper2021highly}, but also similar protein-related papers which made headlines in both Nature Methods \citep[MaSIF]{gainza2020deciphering} and Nature Machine Intelligence \citep{mendez2021geometric}. It seems that protein folding, protein design, and protein binding prediction \citep{stark2022equibind} all appear to be an extremely potent area of attack for geometric GNNs; just one of many solid reasons why the field of structural biology would benefit from these recent developments \citep{bouatta2021protein}.

\bibliographystyle{elsarticle-harv} 
\bibliography{library}

\end{document}